\documentclass[letterpaper, 10 pt, conference]{ieeeconf} 

\IEEEoverridecommandlockouts 
\usepackage{amsfonts}
\usepackage{amsmath}
\usepackage{float}
\usepackage{color}
\usepackage{tikz}
\usepackage{subcaption}
\usepackage{multicol}
\usepackage{float}
\usepackage[]{algorithm2e}
\usepackage{upgreek}
\usepackage{epsf}
\usepackage{bm}
\usepackage{amsmath} 
\usepackage{amssymb}
 \usepackage{epsfig}
 \usepackage{graphicx}  
 \usepackage{psfrag}
\usepackage{lmodern}

\usepackage{graphicx,psfrag}	
\usepackage[latin1]{inputenc}
\graphicspath{{./figures/}}
\usepackage{url} 

\title{\LARGE \bf
From NLVO to NAO: Reactive  Robot Navigation using Velocity and Acceleration Obstacles
}

\author{Asher Stern and Zvi Shiller$^{1}$
\thanks{$^{1}$ Mechanical Engineering and Mechatronics, Ariel University, Israel. 
{\tt\small shiller@ariel.ac.il}}}%

\begin{document}
\maketitle
\pagestyle{empty}

\begin{abstract}
This paper introduces a novel approach for robot navigation in challenging dynamic environments.  The proposed method builds upon the concept of Velocity Obstacles (VO) that was later extended to Nonlinear Velocity Obstacles (NLVO) to account for obstacles moving along nonlinear trajectories. The NLVO is extended in this paper to  Acceleration Obstacles (AO) and Nonlinear Acceleration Obstacles (NAO) that account for velocity and acceleration constraints.  Multi-robot navigation is achieved by using the same avoidance algorithm by all robots.  At each time step, the trajectories of all robots are predicted based on their current velocity and acceleration to allow the computation of their respective NLVO, AO and NAO.    

The introduction of AO and NAO allows the generation of safe avoidance maneuvers that account for the robot dynamic constraints better than could be done with the NLVO alone.  This paper demonstrates the use of AO and NAO for robot navigation in challenging environments.  It is shown that using AO and NAO enables simultaneous real-time collision avoidance while accounting for robot kinematic and a direct consideration of its dynamic constraints.  The presented approach enables reactive and efficient navigation, with potential application for autonomous vehicles operating in complex dynamic environments. 
\end{abstract}

\section{Introduction}
Dynamic environments represent an important and growing segment of modern
automation with applications as diverse as, ground,
aerial and marine autonomous vehicles (AV), air and sea traffic control,
automated wheelchairs and even virtual animation and virtual reality games.
Common to these applications is the need of a decision system able to quickly select maneuvers that avoid potential collisions with static and moving obstacles, while moving towards a specified goal, or track a predefined path.  The challenge of
such a decision system is collision avoidance with any number of static and moving obstacles and reaching the goal, while considering robot dynamics and the future trajectories, known or estimated, of the moving obstacles.
This is a serious challenge since the connectivity of the configuration space in dynamic environments, and hence the goal's reachability may change during motion. 
The objectives of the planner in dynamic environments are therefore prioritized to first ensure the survival of the robot because reaching the goal is not guaranteed since the connectivity of the free configuration space over time cannot be guaranteed.       

An effective approach to avoid collisions in dynamic environments is the use of the Velocity Obstacle (VO) \cite{ref:fiorini-7} that maps obstacles, static or dynamic, to the velocity space of the maneuvering robot.  The  Velocity Obstacle (VO), represents the set of colliding velocities between the robot and an individual obstacle.  Selecting a velocity outside the VO of all obstacles ensures collision-free motion while the obstacles are moving at constant velocities.  The VO was extended to the Nonlinear Velocity Obstacle (NLVO), which accounts for arbitrary known or predicted trajectories of the obstacle \cite{Shiller-2001}. It allows much fewer velocity adjustments than the linear version \cite{ref:fiorini-7} when the obstacle is moving along curved trajectories.  

Another variant of the VO is the Reciprocal VO (RVO) \cite{reciprocal2008} \cite{bareiss2015generalized}. It assumes multi-robot avoidance where each robot is expected to contribute to the avoidance effort.  Geometrically, the RVO is a scaled-down version of the original VO so that each robot makes only a partial effort to avoid the other obstacle (by avoiding a smaller VO), letting the other robot reciprocate by sharing the mutual avoidance maneuver.  It was claimed that the use of RVO avoids oscillations that were attributed to the original VO \cite{reciprocal2008}. This is a misleading statement since the oscillations are caused by inconsistent selection of the avoiding velocity.  Furthermore, the full avoidance of the VO at each time step is not necessary and often not possible due to the acceleration constraints that are often ignored in kinematically oriented algorithms. It is about time to refute this misleading characterization of the VO paradigm.    
A comprehensive review of current literature on motion planning using the Velocity Obstacle paradigm can be seen in  \cite{VESENTINI2024104645}.

In this paper, we address the obstacle avoidance problem in the acceleration domain by extending the VO to AO (Acceleration Obstacle) and the NLVO to NAO (Nonlinear Acceleration Obstacle).  This is motivated by the fact that a robot moving in a dynamic environment is a dynamic and not a kinematic system.  The simplest model for such a system is of  second order that is driven by  acceleration that can be arbitrarily selected subject to the robot's  acceleration constraints.    

The Acceleration Obstacle, AO, in analogy to the Velocity Obstacle, VO, consists of the constant {\it accelerations} that would cause collisions between a robot and a moving obstacle.  Unlike the VO, the geometric shape of the AO depends on the initial velocities of the robot and the obstacle and the obstacle's initial acceleration, as will be shown later in this paper.    

The AO was earlier addressed in \cite{van2011reciprocal}, and more recently in \cite{AO2022} in the context of navigation in human crowds. 
Although originally conceived in \cite{van2011reciprocal}, the AO was not used then for the reason that accelerations tend to change frequently and are therefore difficult to observe.  To this end, they proposed the Acceleration Velocity Obstacle, AVO, which is similar to the VO, except that it accounts for the transition from the current to the target velocity using a proportional feedback law on the acceleration.  Our experience shows that the accelerations applied by the moving obstacle, during short or long durations, are crucial in selecting the robot's proper avoidance maneuver (it is often sufficient for a short acceleration to divert the obstacle away from a collision course).          
                   
The AO was more recently rigorously introduced in \cite{AO2022} in the context of navigation in human crowds. The AO is derived for a robot and obstacle with an initial relative location and velocity, under the assumption that the obstacle is moving at a constant relative acceleration.  They construct the AO as a union of disks, each expressing the constant relative acceleration that would cause collision between the robot and the obstacle at a specific time.  
It was noted that the shape of the AO is curved, in contrast to the triangular shape of the VO. However, the effect of the initial conditions (velocity and acceleration) on the AO's shape was not rigorously addressed. We later show that the AO is a straight cone when the relative initial velocity is zero.  
We are not aware of other works that explicitly address AO in the context of motion planning.      

If obstacles (or vehicles) are known to
move along general trajectories, then deriving the
nonlinear acceleration obstacle (NAO) to reflect their exact motion can result in fewer
adjustments by the avoiding vehicle compared to using the AO.  In addition, the NAO, may be essential in cases where the AO indicates
a collision, when in fact there is none because of the vehicle's curved motion.  For example, an obstacle moving on the inner lane of a curved road may appear to be on a collision course with the vehicles on the outer lane, using the AO, when in fact there is no imminent
collision if the obstacle stays on the curved lane.  The NAO that takes into account road geometry should eliminate this confusion.  Another example is the problem of merging with traffic along a turn-around, where the vehicles moving around the curve appear to be on a collision course with the vehicle waiting to join.  Here too, the NAO will allow the waiting vehicle to plan an efficient maneuver that safely merges with the traffic flow.

This paper first reviews the derivation of the VO and NLVO and then extends both to the Acceleration Obstacle (AO) and  the Nonlinear Acceleration Obstacle (NAO), which represent the set of constant accelerations applied by the maneuvering robot that would lead to collisions with obstacles moving at constant accelerations and along arbitrary trajectories, respectively. Considering arbitrary obstacle trajectories greatly simplifies the avoidance process, as it requires fewer acceleration adjustments by the maneuvering robot than under the assumption of constant acceleration, as is shown in this paper.  

\section{The Velocity Obstacle}

The velocity obstacle represents 
the set of forbidden velocity vectors at a given time that would cause collision between a robot and an obstacle (static or moving) \cite{ref:fiorini-7}. 
The geometry of this set 
can be easily described in the configuration space, as discussed below.  For simplicity,
we consider planar robots and obstacles.    
The robot and obstacles can be of general shapes, however, to reduce the dimensionality of the problem, we assume a circular robot.  Growing the obstacles by the
radius of the robot transforms the problem into the configuration space where a point robot avoids
circular obstacles in the plane.  It is assumed that the instantaneous states
(position and velocity) of the obstacles are either known or measurable.

A few words about notation: henceforth, $A$ denotes a point robot, 
$\cal{B}$ denotes the set of points defining the geometry of an
obstacle relative to a body fixed frame.  Since the obstacle is solid, $\cal{B}$ does not depend on
$t$. The configuration of the robot is $c(t)\in R^2 , t\in (t_0, t_h)$, defined relative to an inertial frame, where $c(t)$ represents the location of the obstacle center along a general trajectory.  

$B(t)\subset R^2$ denotes the set of points occupied by the obstacle at time
$t$.  It can be expressed as $B(t) = c(t) \oplus \cal{B}$, where $\oplus$ is the Minkowski sum.  

We denote a {\it ray} by $a/b$, consisting of the half line that originates at $a$, passes by $b$, and
does not include $a$.  Similarly, $a/v$ denotes a ray that is parallel to $v$ and originates at $a$.

\subsection{The Linear V-Obstacle}

\begin{figure}[ht]
\vspace{.1cm}
  \centerline{\resizebox{9cm}{!}{\includegraphics{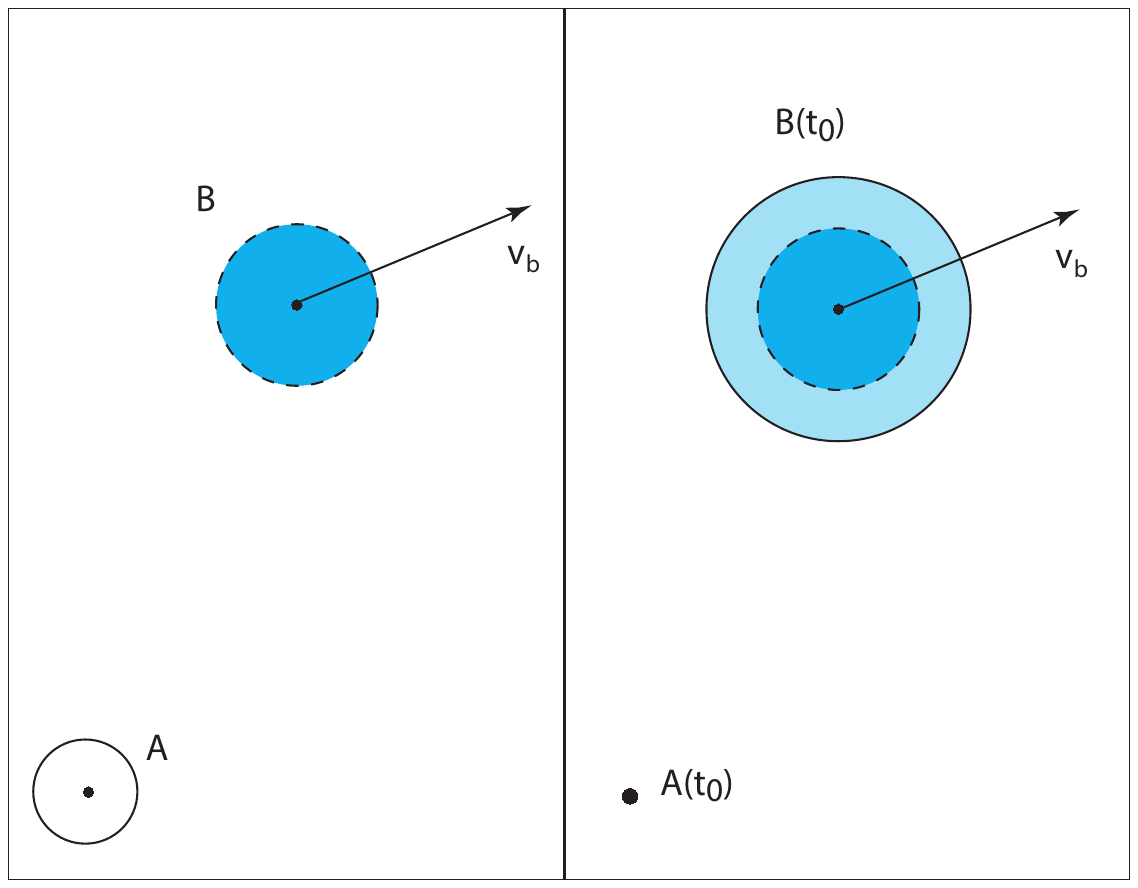}}}
   \caption{A moving obstacle}
   \label{scenario}
 \end{figure}

The linear v-obstacle is demonstrated for the scenario shown in 
Figure~\ref{scenario}, where, at time $t_0$, obstacle $B$ is moving  at some linear velocity $v_b$.

We wish to characterize at time $t=t_0$ the set of velocities of the circular robot $A$ that would cause collision with $\cal{B}$ at any time $t>t_0$. 
We first reduce $A$ to a point and enlarge $B$ by the radius of $A$ to transform the problem into a point colliding with a generally shaped translating object.    

The linear v-obstacle at time $t_0$ is constructed by first generating  
the so called {\it relative velocity cone} ($RVC$).  This is done by sweeping a half line from $A$ along $\partial B(t_0)$, the boundary of $B(t_0)$.  $RVC$ is thus defined as the union of all rays originating from $A$ and passing through $\partial B(t_0)$:  
\begin{equation} 
RVC = \cup A/b, \;\; b \in \partial B(t_0).  
\end{equation}  
The set $RVC\subset R^2$ is the set of all velocities of $A$ relative
to $B$, $v_{a/b} \not=0$, that would result in collision at some time $t \in (0,\infty)$,
assuming that the obstacle stays on its current course at its current
speed \cite{ref:fiorini-7, ref:ghose}.  Consequently, relative velocities outside of $RVC$ ensure avoidance of $B(t)$ at all times $t \in (0,\infty)$ under the same assumptions.  It follows that velocities on the boundary of $RVC$ result in $A$ grazing $B(t)$ at some time $t>t_0$.  The  time of contact depends on the magnitude of $v_{a/b}$.   

\begin{figure}[ht]
 \vspace{.1cm}
\centerline{\resizebox{9cm}{!}{\includegraphics{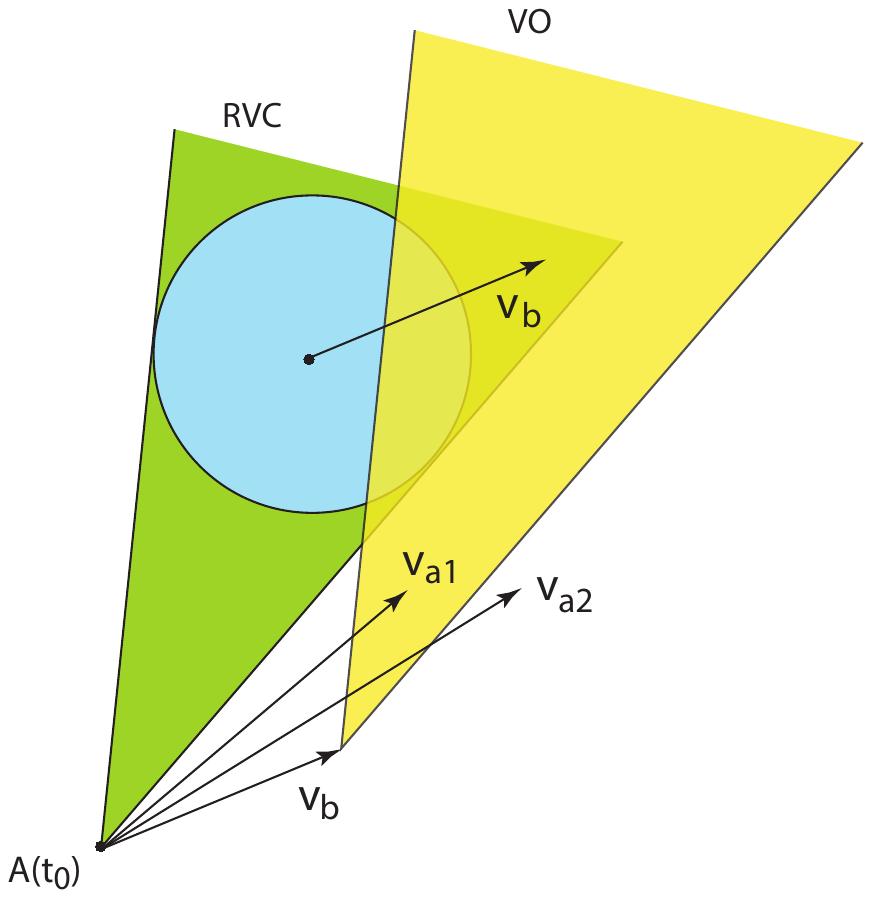}}}
\caption{The Linear Velocity Obstacle}
\label{fig:lv0}
\end{figure}

Translating $RVC$ by $v_b$ produces the \emph{velocity obstacle}, $VO\subset \mathbb{R}^2$ \cite{ref:fiorini-7} shown in Fig. \ref{fig:lv0}: 
\begin{equation} 
VO = v_b \oplus RVC.  
\end{equation}   
Thus, VO represents a
set of {\it absolute} velocities of $A$, $v_a$, that would result in
collision at some time $t \in (0,\infty)$.  Geometrically, each point $v\in VO$ represents a velocity vector in a coordinate system centered at $A(t_0)$.  In Figure~\ref{fig:lv0}, $v_{a1}$ is a colliding velocity,
whereas $v_{a2}$ is not.  Points on the boundary of VO represent absolute velocities $v_a$ that would cause $A$ to graze $B$.  Note that VO is independent of $v_a$, since it reflects the positions of $A$ and $B$ at $t=t_0$, and the obstacle velocity, $v_b$. We call VO a {\it linear} v-obstacle because it represents the forbidden velocities of an obstacle that moves at a constant velocity along a linear trajectory.  
 
\subsection{Temporal Velocity Obstacle}
The VO represents the velocities of $A$ that would result in collision at any time $t=(0,\infty)$.  It is useful to identify a subset of VO that would
result in collision between $A$ and $B(t)$ at a specific time $t$.  

 \begin{figure}[ht]
 \vspace{.1cm}
  \centerline{\resizebox{9cm}{!}{\includegraphics{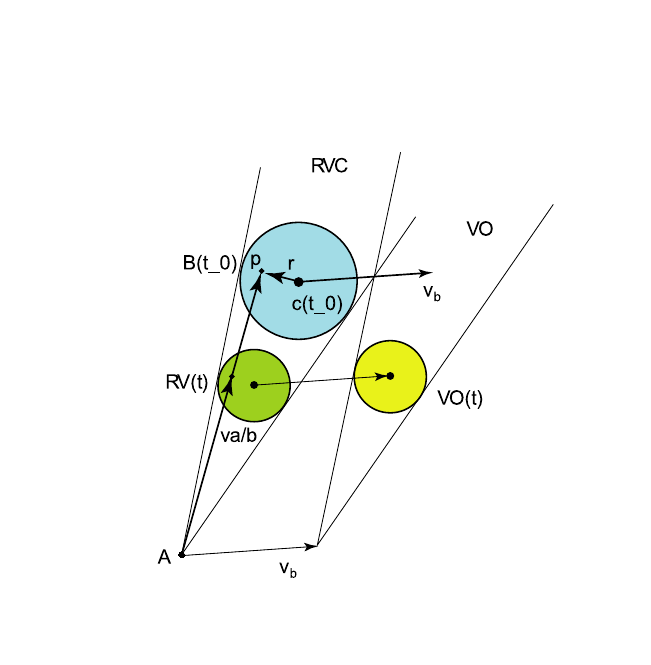}}}
\caption{A Temporal Element of VO.}
\label{fig:tmpvo}
\end{figure}

Consider the relative velocity $v_{a/b} \in RVC$ shown in  Figure~\ref{fig:tmpvo}.   
To reach point $p \in B$ at time $t$, $v_{a/b}$ must satisfy 
 
\begin{equation} 
\label{CT}
v_{a/b}  = \frac{1}{t-t_0}p ,
\end{equation}   
where $p$ is a position vector relative to $A$ (For brevity, we will denote $A(t_0)$ by $A$).  The mapping of $p$ to $v_{a/b}$ (\ref{CT}) is a homothety transformation $H_{A,k}$ 
that maps $A$ to itself and maps any other point $p$ to point $p'$ such that $Ap$ and $Ap'$ are collinear and $Ap = kAp'$.  $A$ is called the {\itshape center} and $k$ the {\itshape ratio} of the homothety \cite{ref:clayton}. 
Here, homothety $H_{A,k}$ scales the vector $p$ by  $k=\frac{1}{t-t_0}$ and positions it at $A$: 
\begin{equation} 
\label{CT1}
v_{a/b}  = H_{A,k}(p); k=\frac{1}{t-t_0}.
\end{equation}  
Substituting for $p$ in (\ref{CT1}) the entire set $B(t_0)$, yields the set of all relative velocities, in a frame centered at $A$, that would result in collision with any point of $B$ at a specific time $t>t_0$, as shown in Figure~\ref{fig:tmpvo}:
\begin{equation} 
\label{RVt}
RV(t) = H_{A,k}(B(t_0)); k=\frac{1}{t-t_0}.
\end{equation}  
To emphasize that the shape of $RV(t)$ is a scaled  $ B$, located at a distance from $A$ that is inversely proportional to the collision time $t$, we represent $p$ by
\begin{equation} 
p = c(t_0) + r
\end{equation}  
where $c(t_0)$ is the position of the center of $ B$ at time $t_0$ relative to $A$, and $r$ is the position of $p$ relative to $c(t_0)$. The set $RV(t)$ then becomes 
\begin{equation} 
\label{RVt2}
RV(t) = H_{A,k}(c(t_0))+H_{c(t_0),k}(B); k=\frac{1}{t-t_0} 
\end{equation}     
or
\begin{equation} 
\label{RVC0}
RV(t) = \{v| v = \frac{c(t_0) + r}{t-t_0}, r \in B\}.
\end{equation}  
Translating $RV(t)$ by $v_b$ produces the set VO($t$), shown in Figure~\ref{fig:tmpvo}, of all {\it absolute} velocities of $A$ that would result in collision with any point of
 $B$ at time $t>t_0$: 
\begin{equation}  
 VO(t) = v_b \oplus RV(t).  
\end{equation}   
 This leads
 to the following formal definition of the linear v-obstacle, VO:
 
\noindent {\bf Definition 1}: The Linear Velocity Obstacle \\
Consider at time $t_0$ a point robot $A$,
located at the origin, and an obstacle $B$ centered at $c(t_0)$
and moving at a constant velocity $v_b$.  The linear v-obstacle, VO, consists of the
set of all linear velocities of $A$ at time $t=t_0$ that would collide with $ B$ at any time $t>t_0$:
\begin{equation} 
\label{vo} \
VO = v_b \oplus \bigcup_{t} H_{A,k}(B(t_0)); k=\frac{1}{t-t_0}; \forall t > t_0.  
\end{equation}

\begin{figure}[ht]
\vspace{.1cm}
  \centerline{\resizebox{8cm}{!}{\includegraphics{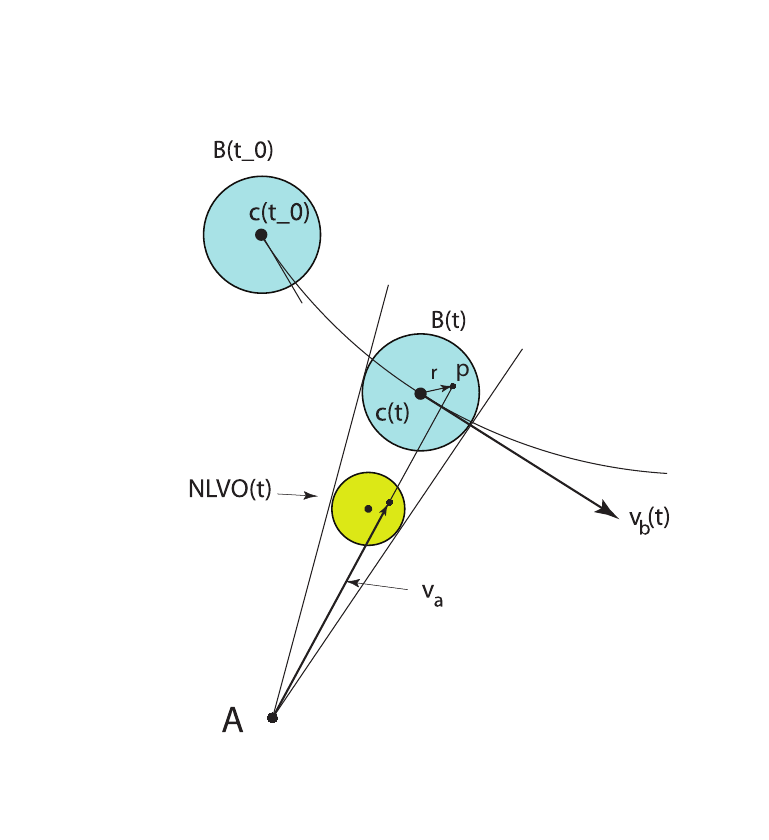}}}
   \caption{A temporal element of the non-linear v-obstacle.}
   \label{fig:nl_v-obst}
 \end{figure}

\subsection{The Non-Linear V-Obstacle}
The non-linear v-obstacle (NLVO) applies to the scenario shown in
Figure~\ref{fig:nl_v-obst}, where, at time $t_0$, a point robot, $A$,
attempts to avoid an obstacle, $ B$, that at time $t_0$
is located at $c(t_0)$, and is following a general known trajectory,
$c(t)$.  The NLVO thus consists of all velocities of $A$ at $t_0$ that would
result in collision with the obstacle at any time $t>t_0$.  Selecting a
{\it single} velocity, $v_a$, at time $t=t_0$ outside the NLVO thus guarantees to avoid collision at all times, or  
\begin{equation} 
(A(t_0)+v_a t) \cap (c(t)+B)=0
\ ; \ \forall t > t_0. 
\end{equation}   
The non-linear v-obstacle is constructed as a union of its temporal elements, NLVO($t$), which is the set of all absolute
velocities of $A$, $v_a$, that would result in collision at a specific time
$t$.  Referring to Figure~\ref{fig:nl_v-obst}, $v_a$ that would
result in collision with point $p \in B(t)$ at time $t>t_0$, expressed in a
frame centered at $A(t_0)$, is simply 
\begin{equation} 
v_a = \frac{c(t)+r}{t-t_0}, 
\end{equation}
where $r$ is the vector to point $p$ in the obstacle's fixed frame.

It is again a homothety transformation, centered at $A(t_0)$ and scaled by $k=\frac{1}{t-t_0}$: 
\begin{equation}  
v_a  = H_{A,k}(c(t)+r); k=\frac{1}{t-t_0}.
\end{equation} 

The set, NLVO($t$) of all absolute velocities of $A$ that would
result in collision with any point in $B(t)$ at time $t>t_0$ is thus: 
\begin{equation} 
\label{nlvot}
NLVO(t) = H_{A,k}(B(t)), \;\;k=\frac{1}{t-t_0}.
\end{equation}
We can rewrite (\ref{nlvot}) using the Minkowski sum to emphasize the geometric shape of NLVO($t$):
\begin{equation} 
NLVO(t) = \frac{c(t)}{t-t_0}\oplus\frac{ B(t)}{t-t_0}, \;\; \forall t > t_0.  
\end{equation}
Clearly, NLVO($t$) is a
scaled $B$, bounded by the cone formed between $A$ and $B(t)$, and located at a distance $\frac{c(t)}{t-t_0}$ from $A$. 
Note that the tangency points of the extreme rays of this cone and $B(t)$ are not necessarily the points
where $A$ grazes $B(t)$, as discussed later.  Note also that the NLVO($t$) is
independent of $v_b(t)$, since it applies only to $B(t)$ and not to its
future or past positions.  This leads to the following formal definition  of the nonlinear v-obstacle: 
    
\noindent {\bf Definition 2}: The Nonlinear Velocity Obstacle \\
Let $A$ be a point robot, located at time
$t=t_0$ at the origin, and $B$ be an obstacle that is moving along a
general trajectory $c(t), t=[t_0,\infty)$.  The non-linear v-obstacle,
NLVO, representing the set of all linear velocities of $A$ that would   collide with $B(t)$ at time $t=(t_0,\infty)$, is defined by 
  \begin{equation} 
\label{nlvo}
NLVO = \bigcup_{t}H_{A,k}(B(t)); k=\frac{1}{t-t_0}; \forall t > t_0. 
\end{equation}

\begin{figure}[ht]
\vspace{.1cm}
  \centerline{\resizebox{6cm}{!}{\includegraphics{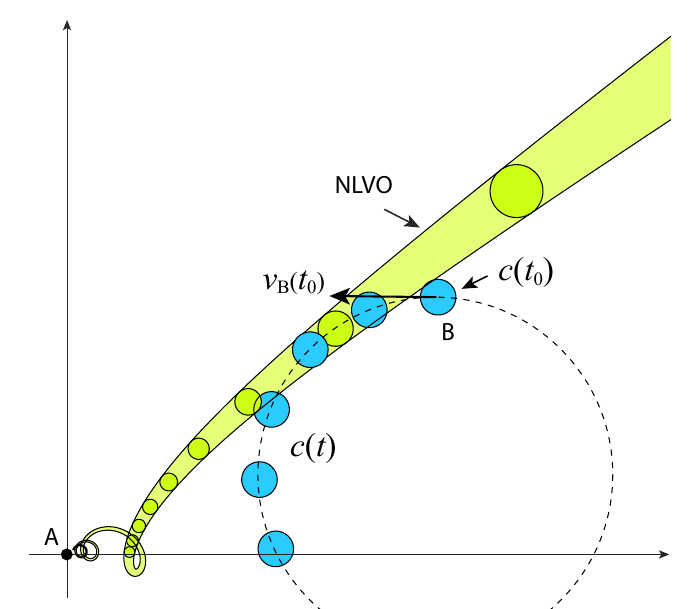}}}
   \caption{Construction of the  NLVO for obstacle B moving along a circular path $c(t)$ at a constant tangent velocity $v(t_0)$.  For a bounded path, the NLVO is a warped cone terminating at $A(t_0)$.}
   \label{fig:nlvo}
 \end{figure}

 The non-linear v-obstacle is a warped cone as shown in Figure~\ref{fig:nlvo}.  If $c(t)$ is bounded, then the apex of this cone is at $A(t_0)$.  The NLVO can be generated for graphical simulations by drawing its individual temporal elements at discrete time intervals, or by drawing its boundaries, which represent velocities that would result in $A$ grazing $B$. 

\begin{figure}[h]
\vspace{.1cm}
  \centerline{\resizebox{6cm}{!}{\includegraphics{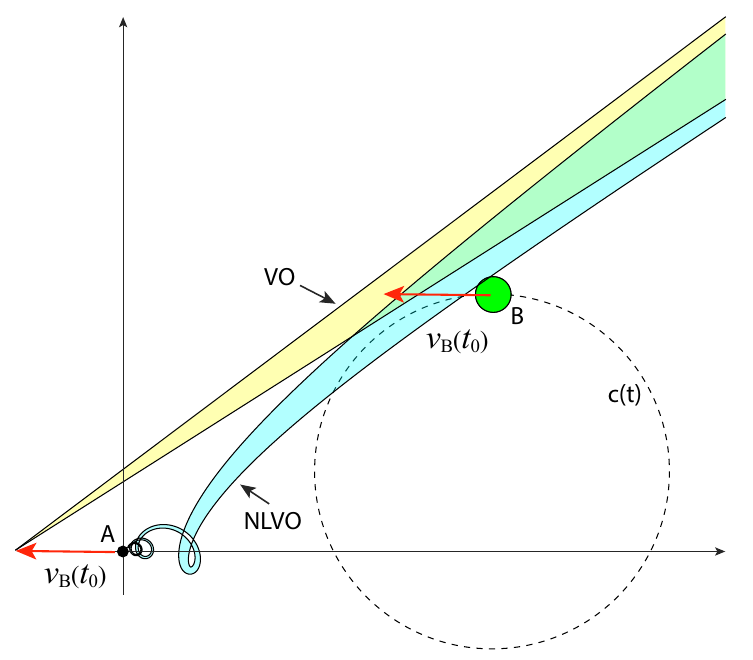}}}
   \caption{A comparison between VO and NLVO for an obstacle that is moving along a curved trajectory $c(t)$.  The NLVO terminates at   $A(t_0)$, whereas the VO that accounts only for the initial velocity at time $t_0$ is shifted by $v(t_0)$ from  $A(t_0)$.}
   \label{fig:vo_nlvo}
 \end{figure}

 \section{Acceleration Obstacle (AO)}
\begin{figure}[ht]
\centerline{\resizebox{8cm}{!}{\includegraphics{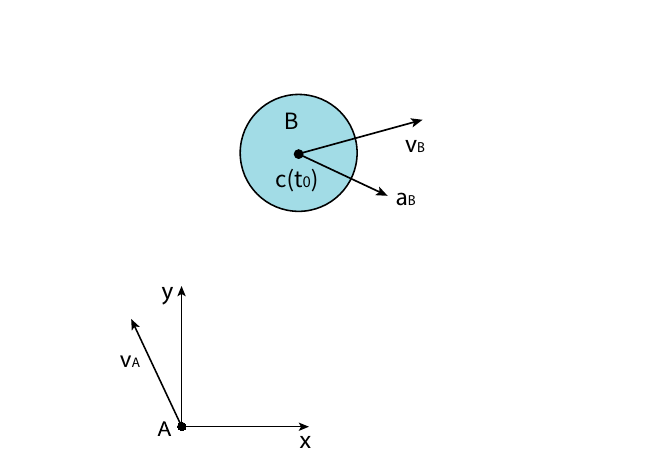}}}
 \caption{A moving robot $A$, and obstacle $B$ that is centered initially at $c(t_0)$ and moving at an initial velocity $v_B$ and a constant acceleration $a_B$.}
   \label{initial_ao}
 \end{figure}

Following the representation of robot and obstacles in the configuration space (see Fig. \ref{scenario}), we start with the case of a circular obstacle $B$, initially located at $c(t_0)$ in a coordinate frame centered at $A$, and moving at a constant acceleration $a_B$ and an initial velocity $v_{B}(t_0)$, as shown in Fig. \ref{initial_ao}.  
We wish to compute the constant acceleration $a(t)$ of a point robot $A$, also shown in Fig. \ref{initial_ao}, that is moving at an initial velocity $v_A (t_0)$ that will reach $c(t)$ at any time $t> t_0$:   
\begin{equation}
c(t_0)+v_B(t_0)t +\frac{1}{2}a_B t^2  =v_{A}(t_0)t+\frac{1}{2}a_At^2
\label{traj_ao}
\end{equation}
Solving for $a_A(t_0)$ yields: 
\begin{equation}
a_A=  \frac{2c(t_0)}{t^2}-\frac{2v_{A/B}(t_0)}{t}+a_B
\end{equation}
Accounting for all points of $B$ defines the set AO($t$) of all constant accelerations that would collide with $B(t)$ at time $t$:
\begin{equation}
AO(t)=  \frac{2B}{t^2}\oplus(\frac{2c(t_0)}{t^2}-\frac{2v_{A/B}(t_0)}{t})+a_B
\label{aot}
\end{equation}

Integrating \eqref{aot} over time produces the set $AO\subset \mathbb{R}^2$ of all constant accelerations that would collide with obstacle $B$ at any time $t>t_0$: 
\begin{equation}
\label{ao}
AO = \bigcup_{t}AO(t);\;\; \forall t > t_0 
\end{equation}This leads
 to the following formal definition of the Acceleration Obstacle AO:
 
\noindent {\bf Definition 3}: The Acceleration Obstacle (AO)\\
Consider at time $t_0$ a point robot $A$,
located at the origin and moving at an initial velocity $v_A(t_0)$, and an obstacle $B$ centered at $c(t_0)$
and moving at an initial velocity $v_B$ and a constant acceleration $a_B$.  The Acceleration obstacle, AO, consists  of the 
set of all accelerations   of $A$ at time $t=t_0$ that would collide with $ B$ at any time $t>t_0$:
\begin{equation}
AO = \bigcup_{t}\{\frac{2B}{t^2}\oplus(\frac{2c(t_0)}{t^2}-\frac{2v_{A/B}(t_0)}{t})+a_B\}; \forall t > t_0
\end{equation}
\begin{figure}[ht]
\centerline{\resizebox{6cm}{!}{\includegraphics{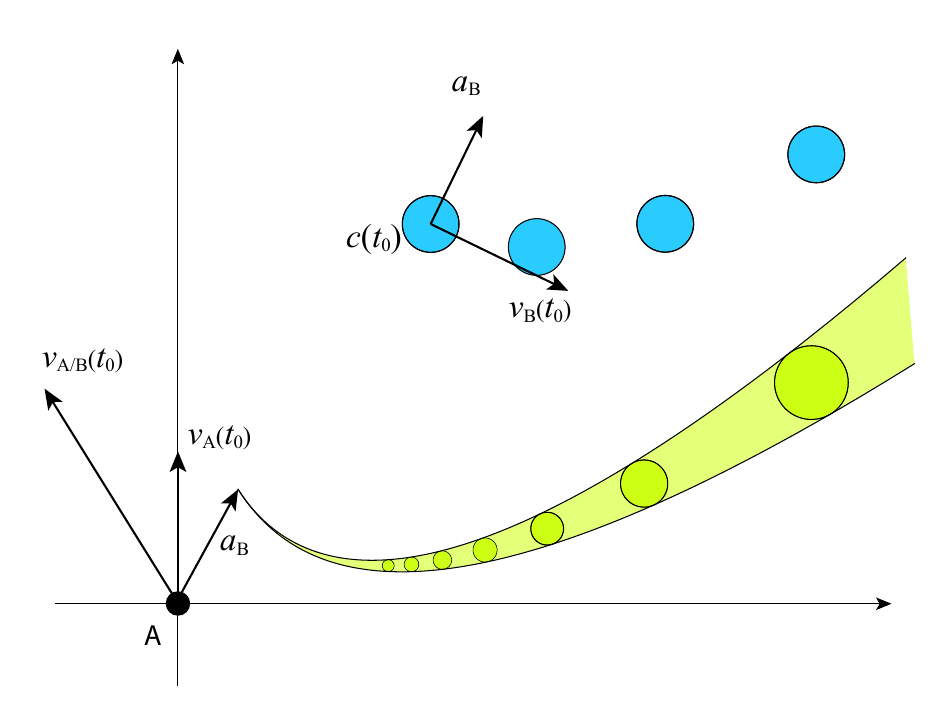}}} \caption{Constructing the AO for $B$ moving at a constant acceleration with an initial velocity $v_B(t_0)$, and $A$ moving at an initial velocity $v_A(t_0)$.}
   \label{tempao}
 \end{figure} %
It is a warped cone, shifted from $A$ by $a_B$ and tangent  to $v_{A/B}(t_0)$, as $t \rightarrow  \infty$, 
as shown in Fig. \ref{tempao}.  

\section{Nonlinear Acceleration Obstacle (NAO)}
The Nonlinear Acceleration Obstacle, NAO, accounts for obstacles that are moving on general (nonlinear) trajectories.

\begin{figure}[ht]
\centerline{\resizebox{6cm}{!}{\includegraphics{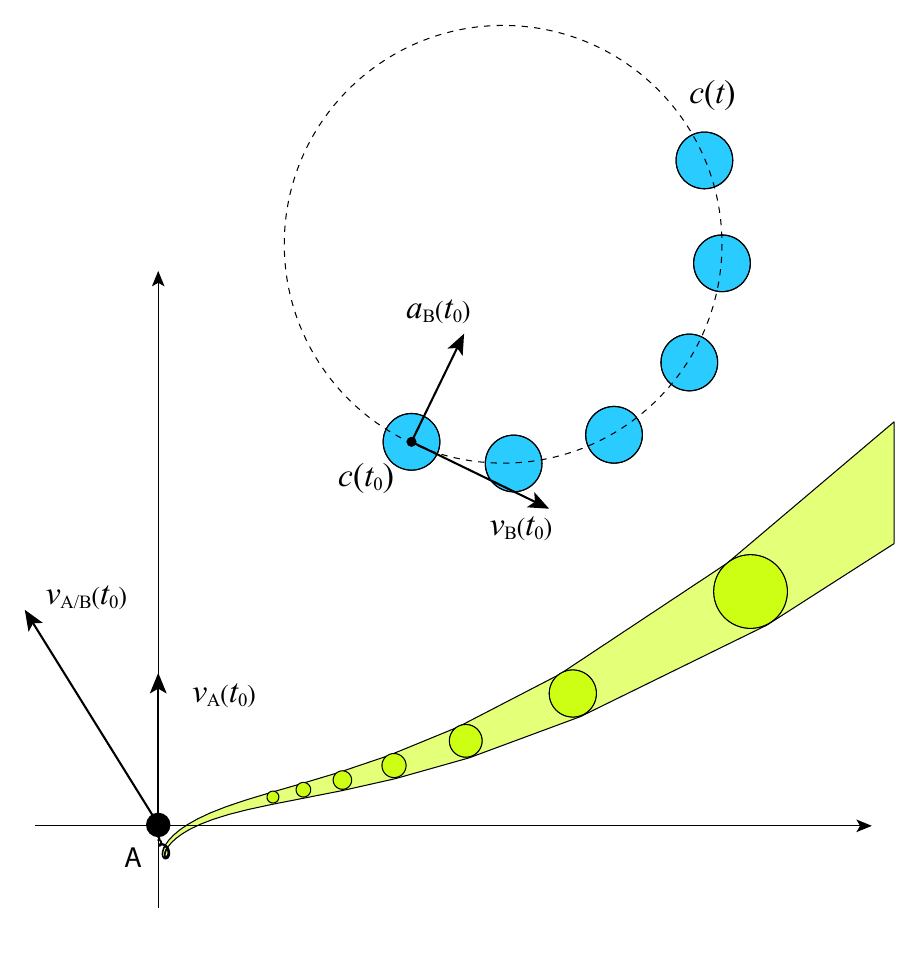}}}
 \caption{Constructing the NAO for $B$ moving along a circular trajectory $B$ at a constant tangent velocity, and $A$ moving at an initial velocity $v_A(t_0)$.}
\label{tempnao}
\end{figure}

The derivation of NAO is similar to the derivation of AO except that the left hand side of Eq. \eqref{traj_ao} is replaced with $c(t)$, the actual location of the obstacle center at time $t$:     
 \begin{equation} 
c(t) =v_{A}(t_0)t+\frac{1}{2}a_A(t_0)t^2.
\label{traj_nao}
\end{equation}  
Solving for $a_A(t_0)$:
\begin{equation}
a_A(t_0)=  \frac{2c(t)}{t^2}-\frac{2v_{A}(t_0)}{t}
\end{equation} 
The temporal Nonlinear Acceleration Obstacle is thus:
\begin{equation}
NAO(t)=  \frac{2B}{t^2} \oplus (\frac{2c(t)}{t^2}-\frac{2v_{A}(t_0)}{t})
\end{equation} 
\noindent {\bf Definition 4}: The Nonlinear Acceleration Obstacle (NAO) \\
Consider at time $t_0$ a point robot $A$,
located at the origin and moving along and an obstacle $B$ centered at $c(t_0)$
and moving along an arbitrary trajectory $c(t), t\in (t_0,t\infty)$.  The Nonlinear Acceleration obstacle, NAO, consists  of the 
set of all accelerations   of $A$ at time $t=t_0$ that would collide with $ B$ at any time $t>t_0$:
\begin{equation}
NAO = \bigcup_{t}\{\frac{2B}{t^2}\oplus(\frac{2c(t)}{t^2}-\frac{2v_{A}(t_0)}{t})\}; \forall t > t_0
\end{equation} 
Fig. \ref{tempnao} shows the NAO for obstacle $B$ that is moving along a circular trajectory  $c(t)$ at a constant tangent velocity $v_B$. Since $c(t)$ is bounded, as $t\rightarrow \infty$, the NAO reduces to a point at the origin $A$.     

\section{Selecting an Avoidance Maneuver}
The AO and NAO can be used to plan avoidance maneuvers in ways similar to what was suggested in \cite{ref:fiorini-7} in the context of Velocity Obstacles (VO), namely checking if the current acceleration of the maneuvering vehicle points to the NAO of any obstacle (static or moving). 
If it does, then a safe acceleration must be selected outside of all respective NAO's, within the set of admissible controls (satisfying the control constraints), if one exists.   This would ensure the avoidance of all obstacles until the next time step when this process would repeat.   
This can be done using AO if the obstacle is moving at a constant acceleration, or using NAO otherwise to decrease the update rate of the host's acceleration.   
  
It is important to note that AO and NAO account for the host's current velocity. Hence, selecting a safe acceleration, if one exists, would generate a safe avoidance maneuver as long as the obstacle maintains its current trajectory.    

We assume the host to be a second order system for which the acceleration serves as the control variable. 
This allows the {\it instantaneous} selection of a safe acceleration and thus an immediate safe response to an imminent collision. It also nullifies the use velocity based avoidance  \cite{ref:fiorini-7, Shiller-2001, van2011reciprocal}.    

The selection of the safe acceleration can be guided by various heuristics, such as reaching the goal as fast as possible, minimizing the deviation from the current acceleration, or maximizing clearance with other obstacles.  This paper has focused on collision avoidance in the plane.  The same approach can be extended to three-dimensional and higher spaces to control autonomous drones and air vehicles.

\section{Examples}
\begin{figure}
\vspace{.1cm}
\centerline{\resizebox{8cm}{!}
{\includegraphics{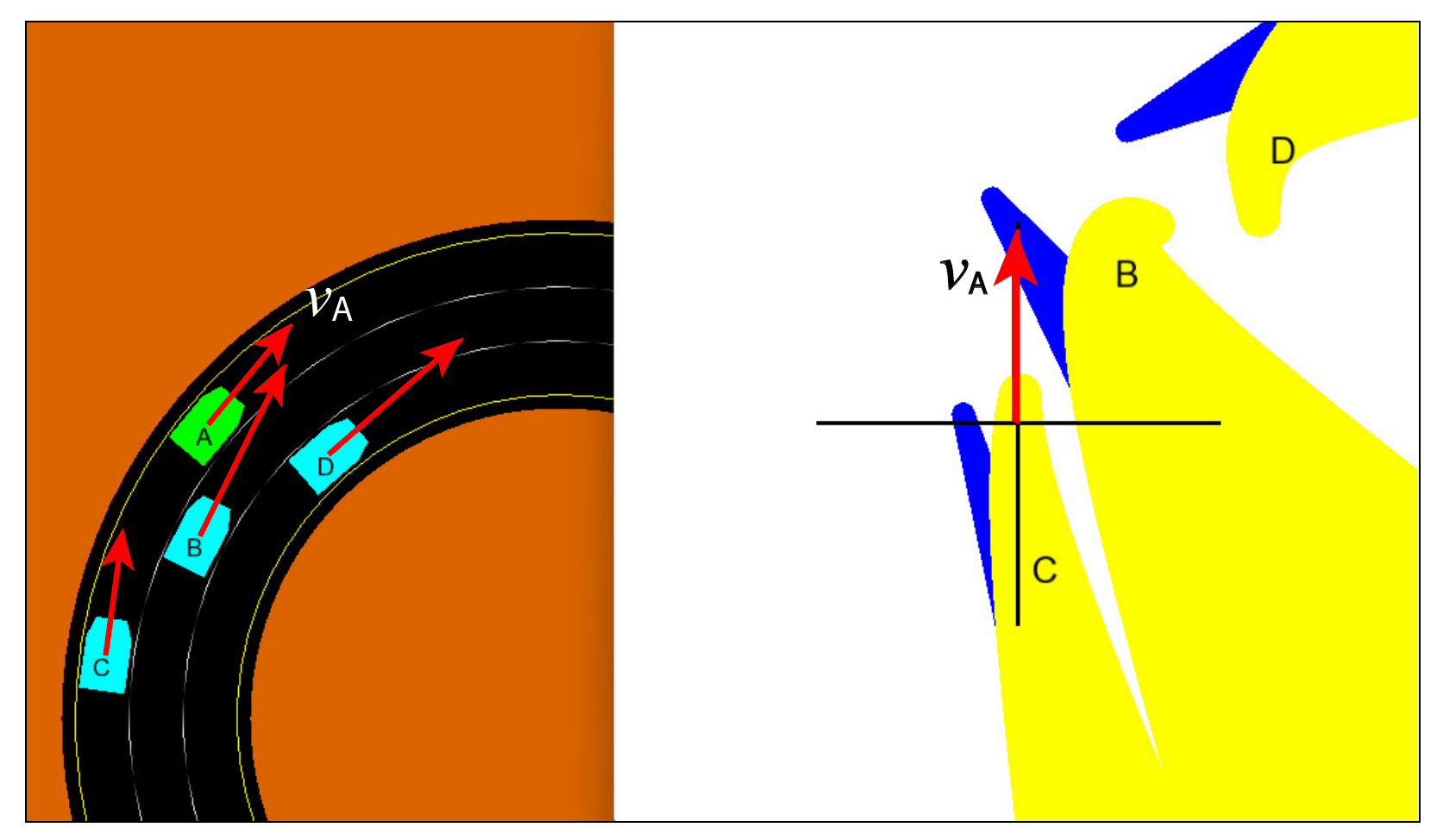}}}
   \caption{Comparing linear and nonlinear v-obstacles along a curved road.}
   \label{fig:ex7}
 \end{figure}

Figure~\ref{fig:ex7} shows four vehicles moving along a curved road. The maneuvering robot is marked $A$, and the other three passive vehicles are $B$, $C$, and $D$. The right-hand side of Figure~\ref{fig:ex7} shows the velocity space of $A$ in its coordinate frame, with the velocity $v_A$ pointing along the vertical axis.  Also shown are the linear (in blue) and non-linear (in yellow) velocity obstacles of $B$, $C$, and $D$.  At the position shown, $v_A$ penetrates the linear velocity obstacle of $B$ (in blue), which implies that $B$ is on a collision course with $A$ had both maintained their current velocities.  Accounting for $B$'s curved trajectory, which is reflected in its nonlinear velocity obstacle (in yellow), shows no potential collision.  Using the linear velocity obstacle would have required $A$ to continuously adjust its velocity to avoid a collision with $B$. In the case shown, it would require $A$ to slow down or speed up.  This, however, is not necessary if $B$ maintains its current course along the middle lane.    
 
\begin{figure}
\vspace{.1cm}
\centerline{\resizebox{8cm}{!}{\includegraphics{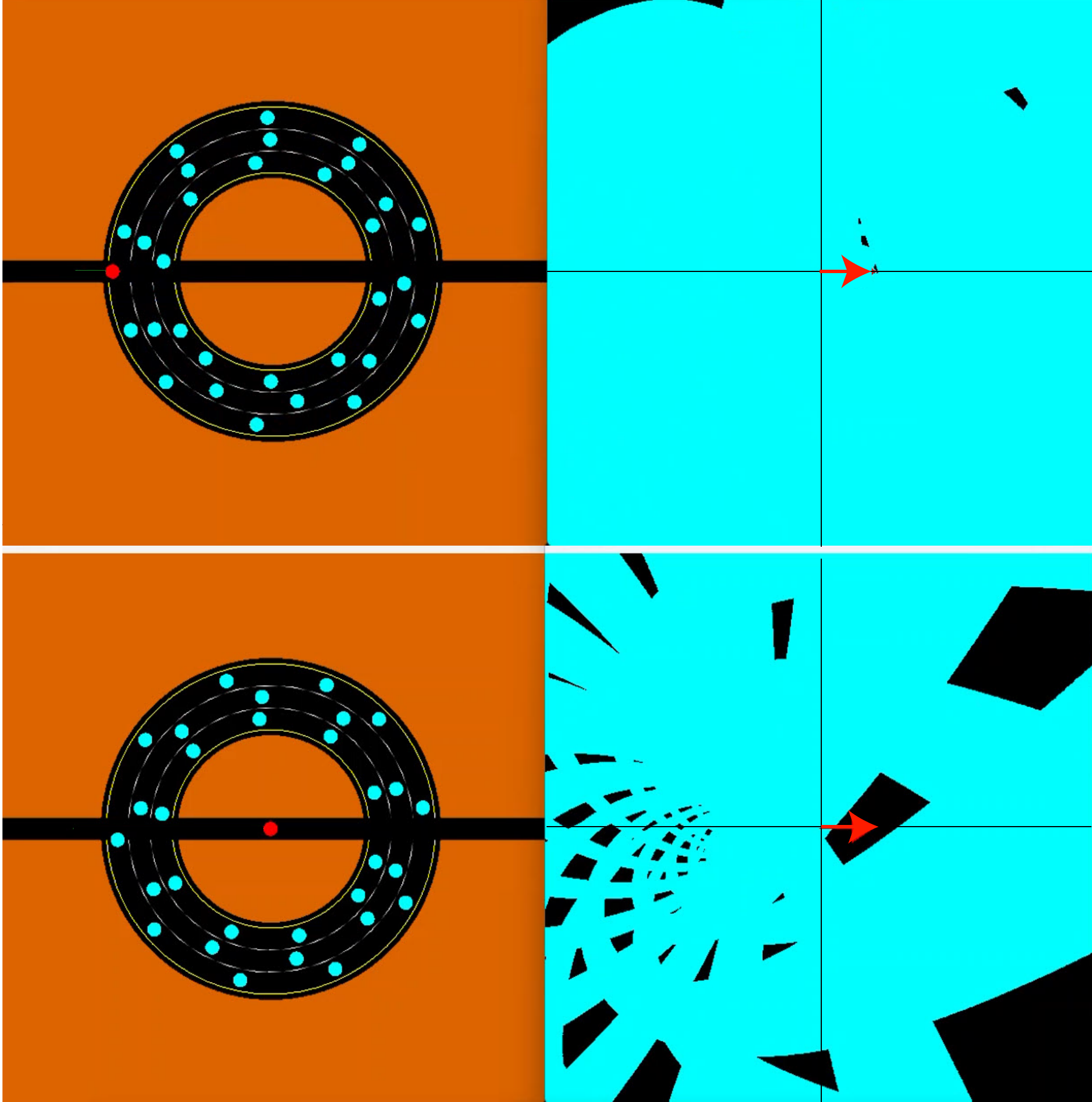}r}
} \caption{A vehicle crosses at a constant acceleration through a circle along which multiple vehicles are moving in three circular lanes.  The horizontal acceleration was selected to point to a collision-free space in the acceleration domain.  The free NAO space is growing as the vehicle faces fewer obstacles to avoid, as shown on the right.}
   \label{naocircle}
 \end{figure} 

 Fig. \ref{naocircle}
shows a vehicle (represented by a red circle) that crosses a busy roundabout with 30 vehicles that are moving in three circular lanes at different speeds in each lane.  
The vehicle is crossing the roundabout at a constant acceleration, selected on the NAO map shown on the right. The selected acceleration is shown as a red arrow on the NAO map.  The vehicle crossed all obstacles with no collision.  Attempting to do the same with AO by frequently adjusting the acceleration to avoid collisions with the nearest vehicles resulted in multiple collisions with the circling vehicles.
 
\begin{figure}
{
\vspace{.1cm}
\centerline{\resizebox{9cm}{!}{\includegraphics{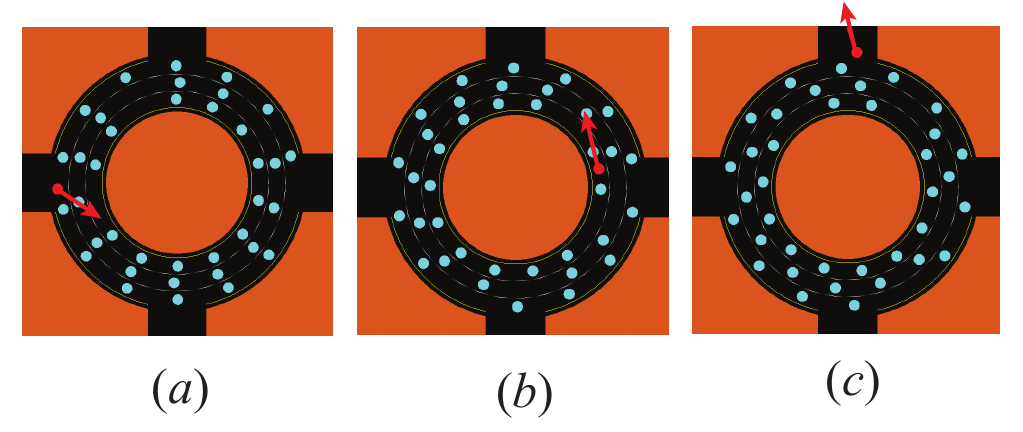}}}
 \caption{Navigating through a busy roundabout using NAO. The maneuvering vehicle, shown in Red, enters the roundabout from the left (a) and navigates through the busy traffic to the top exit (c). 
}   \label{roundnao}}
 \end{figure}

Fig. \ref{roundnao} shows a maneuvering vehicle (represented by a red circle) crossing a busy roundabout with  30 vehicles that are moving in three circular lanes at different speeds in each lane.  The vehicle is maneuvering the roundabout at accelerations selected outside the NAO's computed at specified time intervals. The selected accelerations in Fig. \ref{roundnao} at three positions as a red arrow, attached to a red dot that represents the maneuvering vehicle.  The vehicle exited the roundabout safely with no collision. 

These examples demonstrate the usefulness of the NAO in negotiating complex dynamic environments,  and in locally selecting dynamically feasible collision-avoiding accelerations. The resulting vehicle motions are smooth, resembling the behavior of experienced and careful drivers.

\section{Conclusions}

The original concept of velocity obstacles \cite{ref:fiorini-7} was first generalized to nonlinear velocity obstacles \cite{Shiller-2001} to account for obstacles
moving on arbitrary trajectories.  The nonlinear v-obstacle consists of a
warped cone that is a time-scaled map of the obstacle along its trajectory.
Selecting a single velocity vector outside the nonlinear v-obstacle
guarantees  avoidance of the obstacle during the time interval for which the
v-obstacle was generated.  

The velocity obstacle and non-linear velocity obstacle were extended to acceleration obstacles AO and non-linear acceleration obstacles NAO, respectively.  Both allow the maneuvering vehicle use its acceleration to avoid collisions in complex dynamic environments.   
The AO and NAO were generated similarly to the NLVO  by integrating the temporal AO($t$) and NAO($t$), respectively.    Using NAO allows for
more efficient avoidance  (fewer adjustments) than using AO for obstacles moving along general trajectories.  The result is safer avoidance maneuvers in complex situations, as was demonstrated in several challenging scenarios. Using acceleration to drive the maneuvering vehicle results in smooth motions that account for the vehicle's current velocity and its acceleration constraints.

\bibliographystyle{plain}
\bibliography{main1.bib}
\end{document}